%% file: root.tex
\documentclass[letterpaper, 10 pt, conference]{ieeeconf}  

\input{tex/preamble}

\title{\LARGE \bf Robust and Efficient Embedded Convex Optimization\\through First-Order Adaptive Caching}

\author{Ishaan Mahajan$^{1}$ and Brian Plancher$^{2,3}$
\thanks{This material is based upon work supported by the National Science Foundation (under Award 2411369). Any opinions, findings, conclusions, or recommendations expressed in this material are those of the authors and do not necessarily reflect those of the funding organizations.}%
\thanks{$^{1}$Ishaan Mahajan is with the School of Engineering and Applied Science, Columbia University. {\tt\footnotesize iam2141@columbia.edu}}%
\thanks{$^{2,3}$Brian Plancher is with Barnard College, Columbia University and Dartmouth College. {\tt\footnotesize plancher@dartmouth.edu}}%
}

\begin{document}

\maketitle
\thispagestyle{empty}
\pagestyle{empty}

\begin{abstract}
\input{tex/0_abstract}
\end{abstract}

\section{Introduction}
\input{tex/1_intro}

\section{Background}
\label{sec:background}
\input{tex/2_background}

\section{First-Order Adaptive Caching}
\label{sec:FOAC}
\input{tex/3_adaptive}

\section{Experiments and Results}
\label{sec:ResultsExperiments}
\input{tex/4_experiments}

\section{Conclusion and Future Work}
\label{sec:conc}
\input{tex/5_conclusion}


\section*{Acknowledgements}
The authors would like to thank James Anderson for his insightful comments and feedback. The authors are also indebted to Moises Mata, Emre Adabag, Brandon Pae, Charles Chen, and En Kai (Gino) Zhang for their help with hardware experiments, as well as Gabriel Bravo-Palacios for his support with manuscript preparation and editing. 

\bibliographystyle{IEEEtran}
\bibliography{bib}

\end{document}

%% file: tex/preamble.tex
\IEEEoverridecommandlockouts                              
\usepackage{graphics} 
\usepackage{epsfig} 
\usepackage{mathptmx} 
\usepackage{subcaption}
\usepackage{multirow}
\usepackage{xcolor}

\usepackage{url}
\usepackage{color}
\usepackage{wrapfig}
\newif\iffigs
\figstrue 

\usepackage{float}
\usepackage{setspace}
\usepackage{graphicx}
\usepackage{mathrsfs}
\usepackage{amssymb}
\usepackage{nicefrac}
\usepackage{algorithm}
\usepackage[noend]{algpseudocode}

\makeatletter
\newcommand\fs@spaceruled{\def\@fs@cfont{\bfseries}\let\@fs@capt\floatc@ruled
  \def\@fs@pre{\vspace{0.4\baselineskip}\hrule height.8pt depth0pt \kern2pt}%
  \def\@fs@post{\vspace{-0.4\baselineskip}\kern2pt\hrule\relax\vspace{-12pt}}%
  \def\@fs@mid{\kern2pt\hrule\kern2pt}%
  \let\@fs@iftopcapt\iftrue}
\let\NAT@parse\undefined
\makeatother

\usepackage{hyperref}
\hypersetup{
     colorlinks   = true,
     linkcolor    = blue,
     citecolor    = red
}

\usepackage{adjustbox}
\usepackage{listings}
\lstnewenvironment{itemlisting}[1][]
 {%
  \mbox{}
  \vspace*{-\baselineskip}
  \lstset{
    xleftmargin=\leftmargin,
    linewidth=\linewidth,
    #1
  }%
 }
 {}
\usepackage{flushend}
\usepackage{multicol}

\usepackage{hyperref}
\hypersetup{bookmarks=true}

\usepackage{lipsum}

\usepackage{array}

\usepackage{makecell} 

\usepackage{amsmath,amssymb,amsfonts}

\DeclareMathOperator*{\minimize}{minimize}
\DeclareMathOperator{\subjectto}{subject~to}
\DeclareMathOperator*{\argmin}{argmin}

\newcommand\scalemath[2]{\scalebox{#1}{\mbox{\ensuremath{\displaystyle #2}}}}

\newcommand{\norm}[1]{\|#1\|}

%% file: tex/0_abstract.tex
Recent advances in Model Predictive Control (MPC) leveraging a combination of first-order methods, such as the Alternating Direction Method of Multipliers (ADMM), and offline precomputation and caching of select operations, have excitingly enabled real-time MPC on microcontrollers. Unfortunately, these approaches require the use of fixed hyperparameters, limiting their adaptability and overall performance. In this work, we introduce First-Order Adaptive Caching, which precomputes not only select matrix operations but also their sensitivities to hyperparameter variations, enabling online hyperparameter updates without full recomputation of the cache. We demonstrate the effectiveness of our approach on a number of dynamic quadrotor tasks, achieving up to a 63.4\% reduction in ADMM iterations over the use of optimized fixed hyperparameters and approaching 70\% of the performance of a full cache recomputation, while reducing the computational cost from $O(n^3)$ to $O(n^2)$ complexity. This performance enables us to perform figure-eight trajectories on a 27g tiny quadrotor under wind disturbances. We release our implementation open-source for the benefit of the wider robotics community.

%% file: tex/1_intro.tex
Model Predictive Control (MPC) is a state-of-the-art approach for real-time control of autonomous systems~\cite{wensing2023optimization,kuindersma2016optimization,le2024fast,Bao_etal_autonomousMPC23,grandia2023perceptive}. However, its computational complexity makes it difficult to deploy on tiny robots with limited onboard compute capabilities and fast control rates (e.g., tiny quadrotors)~\cite{neuman_tiny_2022,nguyen2021model}. This has led to a number of past works to develop neural or explicit MPC solutions to precompute control laws and overcome these computational challenges~\cite{torrente2021data,drgovna2022differentiable,salzmann2023real}.

\begin{figure}[!t]
    \centering
    \includegraphics[width=\linewidth]{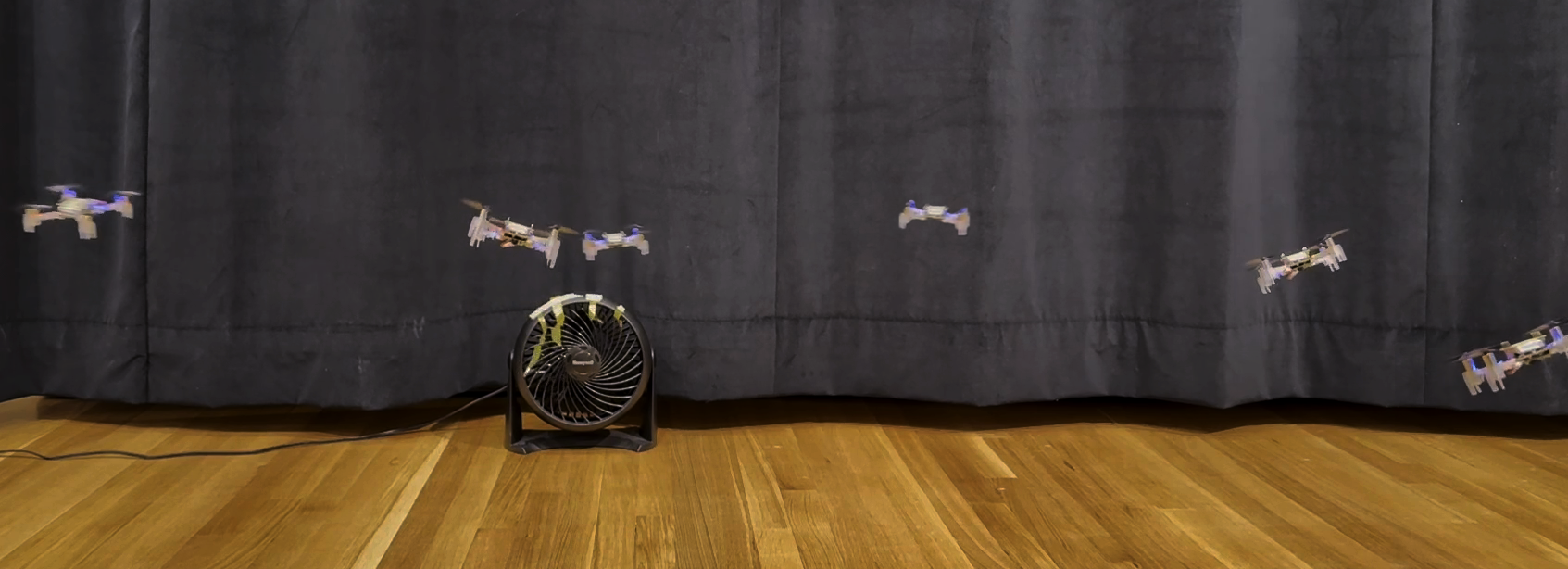}
    \includegraphics[width=\linewidth]{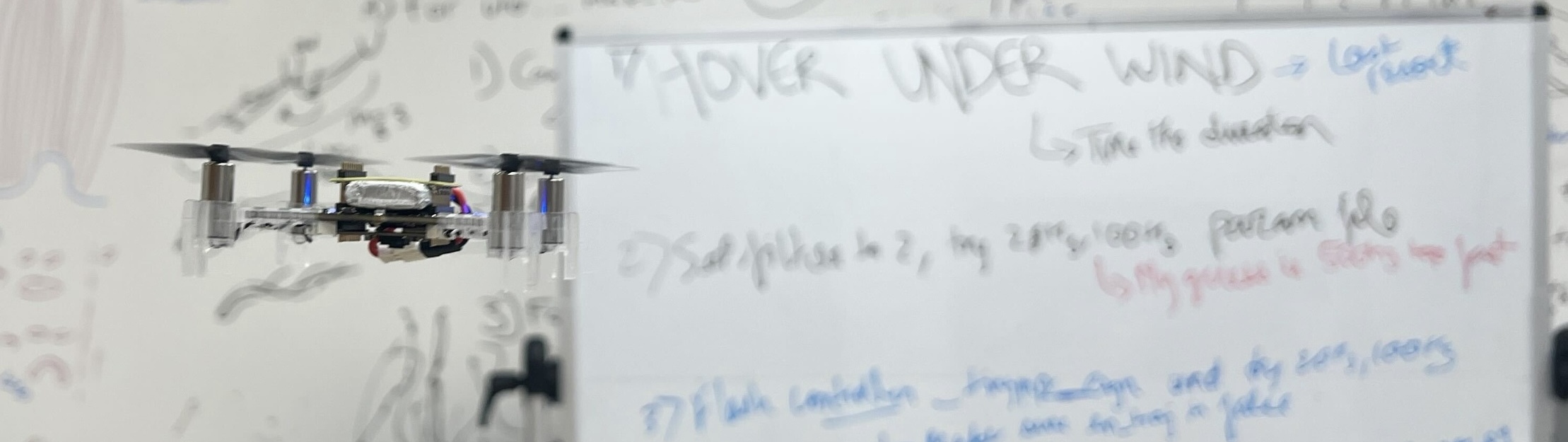}
    \caption{A superimposed image of Crazyflie 2.1+ using First-Order Adaptive Caching under wind disturbances to perform a figure-eight (top) and a stable hover (bottom).}
    \label{fig:hero}
    \vspace{-18pt}
\end{figure}

At the same time, previous work has explored the use of first-order optimization methods, such as the Alternating Direction Method of Multipliers (ADMM), to improve the efficiency of embedded MPC solvers~\cite{boyd2011distributed}. These solvers have demonstrated state-of-the-art performance across a range of hardware platforms, particularly in resource-constrained embedded systems~\cite{verschueren2022acados, ferreau2017embedded, kouzoupis2015first, stellato_osqp_2020, mattingley2012cvxgen}. More recently, techniques that precompute and cache select matrix operations within the ADMM algorithm, leveraging both an offline and online phase, have enabled real-time convex MPC on embedded microcontrollers, enabling dynamic obstacle avoidance on a 27-gram quadrotor~\cite{nguyen2024tinympc, schoedel2024code}. These approaches have been further extended through strategic constraint removal~\cite{hou2024reliablylearn}, model relaxations~\cite{fan2025estrelaxedmodel}, and learned dynamics~\cite{gao2024onchip}, among others. These developments pave the way for a future of highly dynamic, intelligent tiny robots.

Although caching-based methods do significantly reduce computational overhead, and enable real-time performance, they require fixing key hyperparameters in the ADMM solver offline. Unfortunately, hyperparameters such as the ADMM penalty parameter $\rho$ are crucial for ensuring both fast and reliable convergence. In fact, state-of-the-art ADMM solvers typically adapt $\rho$ online through heuristics or learning-based methods to ensure performant operation~\cite{boyd2011distributed, stellato_osqp_2020, he2000alternating, ichnowski2021accelerating}. However, in existing caching approaches, modifying $\rho$ online would require complete cache recomputation, negating the computational benefits of precomputation. In response, previous work~\cite{nguyen2024tinympc, schoedel2024code, bishop2024relu} has attempted to precompute multiple cache sets for different hyperparameters. This approach faces scalability issues due to limited memory onboard embedded compute platforms. 

To address the above limitations, we introduce First-Order Adaptive Caching, a method leveraging first-order sensitivities to efficiently update cached variables in response to changes in key hyperparameters. Inspired by the use of first-order sensitivities to avoid re-training in machine learning based-algorithms~\cite{hose2024parameter}, our approach precomputes both the cache and its sensitivity to hyperparameter variations (e.g., changes in $\rho$). 
This enables a first-order update to the cached values at runtime, \emph{eliminating} the need for full recomputation. In the case of ADMM, this enables a reduction in the computational complexity of cache updates for varying $\rho$ from $O(n^3)$ to $O(n^2)$, while only incurring a $1.5x$ increase in cache size, enabling efficient embedded deployment.

We evaluate our method on a range of dynamic quadrotor tasks, demonstrating a 63.4\% decrease in ADMM iterations as compared to optimized fixed hyperparameters. In fact, our approach is able to achieve 70\% of the performance of a full cache recomputation at the aforementioned much lower complexity. This enhanced efficiency allows a 27g tiny quadrotor to execute maneuvers despite wind disturbances.
To benefit the wider robotics community, we release our implementation as an open-source extension to the TinyMPC software framework, which is available at: \href{https://tinympc.org/}{\texttt{https://tinympc.org}}, providing a scalable and adaptable solution for embedded real-time control.

%% file: tex/2_background.tex
\subsection{Alternating Direction Method of Multipliers (ADMM)}
Consider the generic convex optimization problem:
\begin{subequations}
\begin{align}
\minimize_{x}~& f(x),\label{ADMM_basic}\\
\subjectto~ &x \in \mathcal{C},\label{ADMM_const}
\end{align}
\label{eq:generic_problem}%
\end{subequations}
where $f$ is a twice differentiable cost function and $\mathcal{C}$ is a convex set. ADMM solves~\eqref{eq:generic_problem} by forming the following equivalent problem:
\begin{subequations}
\begin{align}
\minimize_{x, z}~& f(x) + I_{\mathcal{C}}(z),\\
\subjectto~ &x = z,
\end{align}
\label{eq:equiv_problem}%
\end{subequations}
where $z$ is a slack variable and $I$ is an indicator function for the set $\mathcal{C}$ given by:
\begin{equation}
    I := 
    \begin{cases}
    0 ~& if~ z \in \mathcal{C}\\
    \infty ~& otherwise. 
    \end{cases}
    \label{eq:indicator_func}%
\end{equation}
The scaled form of the augmented Lagrangian for this problem is as follows, where $\rho\geq0$ is a penalty parameter, $r=x-z$ is the constraint residual, and $y=\lambda/\rho$ is the scaled dual variable derived from the Lagrange multiplier, $\lambda$:
\begin{equation}
    \scalemath{1.0}{\mathcal{L}_{\rho}(x,z,y) = f(x) + I(z)+\frac{\rho}{2}\norm{r+y}_2^2 - \frac{\rho}{2}\norm{y}_2^2}, \label{eq:mod_Lagrangian}%
\end{equation}
The ADMM algorithm solves this problem through the use of the following three-step iterative process \cite{boyd2011distributed, Bravo_Wensing_ADMM_CD22}:
\begin{subequations}
\begin{align}
    x_{k+1}&:= \argmin_{x}~\mathcal{L}_{\rho}(x,z^k,y^k),\label{ADMM1}\\
    z_{k+1}&:= \argmin_{z}~\mathcal{L}_{\rho}(x_{k+1},z,y_k),\label{ADMM2}\\
    y_{k+1}&:= y_k + \rho r_{k+1}, \label{ADMM3}
\end{align}
\label{eq:admm_steps}%
\end{subequations}
where \eqref{ADMM1} minimizes $x$, \eqref{ADMM2} minimizes $z$, and \eqref{ADMM3}, denoting a sum of residuals, updates $y$. 

In ADMM-based optimization, convergence and stability depend on balancing the primal and dual residuals \cite{boyd2011distributed}, which naturally scale with the magnitude of state deviations:
\begin{equation}
r_k^{\text{prim}} = x_{k} - z_{k}, \quad r_k^{\text{dual}} = -\rho(z_{k}-z_{k-1}).
\label{eq:residuals}
\end{equation}

Adaptive tuning of \(\rho\) is critical to ensure efficient convergence, particularly in dynamic systems where residual magnitudes can vary significantly~\cite{stellato_osqp_2020,Bravo_etal_EngMIwADMM_CD24} as the ADMM penalty parameter \(\rho\) directly influences solver emphasis on primal versus dual feasibility:
\begin{itemize}
    \item Higher \(\rho\) values prioritize primal feasibility (constraint satisfaction).
    \item Lower \(\rho\) values prioritize dual feasibility (optimality).
\end{itemize}

\subsection{ADMM for Convex Model-Predictive Control}
The Model Predictive Control (MPC) problem can be defined as follows:
\begin{subequations}
\begin{align}
    \minimize_{x_{1:N}, u_{1:N-1}}~& J(x_{1:N}, u_{1:N-1}), \\ 
    \subjectto~& x_{k+1} = h(x_k, u_k) \quad \forall k \in [1, N), \label{eq:dynamics} \\
    ~& x_k \in \mathcal{X}, \quad u_k \in \mathcal{U}, 
\end{align}
\label{eq:MPC}
\end{subequations}%
where \(x_k \in \mathbb{R}^n\) and \(u_k \in \mathbb{R}^m\) represent the state and control input at time step \(k\), \(N\) is the prediction horizon, \(h(\cdot)\) defines the system dynamics, and $\mathcal{X}$ and $\mathcal{U}$ are convex sets denoting additional state and input constraints. 

We can choose to separate the minimization of the cost function subject to the dynamics constraints from the remainder of the constraints, $\mathcal{X}$ and $\mathcal{U}$, through the use of ADMM by introducing slack variables $z$ and $w$ associated with the states $x$ and controls $u$, respectively. 
Hence, the optimization problem we solve using ADMM is the following: 
\begin{subequations}
\begin{align}
    \minimize_{x,~u}~& \mathcal{L}_{\rho}(x,u,z,w,y,g) \\ 
    \subjectto~& x_{k+1} = h(x_k, u_k) \quad \forall k \in [1, N), \\
               & x_k = z_k, \quad u_k = w_k,
    \end{align}
\label{eq:admm-MPC}
\end{subequations}
where the augmented Lagrangian $\mathcal{L}_{\rho}(\cdot)$ is given by:
\begin{align}
    \mathcal{L}_{\rho}(x,u,z,w,y,g) ~& = J(x,u) + I_\mathcal{X}(z)+ I_\mathcal{U}(w) \label{eq:cmpc_Lagrangian} \\
    ~& +~\frac{\rho}{2}\sum_{k=1}^N \norm{x_k-z_k+y_k}_2^2 - \frac{\rho}{2}\sum_{k=1}^N \norm{y_k}_2^2 \nonumber \\
    ~& +~\frac{\rho}{2}\sum_{k=1}^{N-1} \norm{u_k-w_k+y_k}_2^2 - \frac{\rho}{2} \sum_{k=1}^{N-1} \norm{g_k}_2^2, \nonumber 
\end{align}
with scaled dual variables \(y_k = \tfrac{\lambda_k}{\rho}\) and \(g_k = \tfrac{\mu_k}{\rho}\), derived from the Lagrange multipliers $\lambda_k$ and  $\mu_k$, respectively.  

The Lagrange multipliers $\lambda_k$ and  $\mu_k$ enforce consensus between \((x_k, z_k)\) and \((u_k, w_k)\), respectively. In more conventional formulations (e.g., \cite{stellato_osqp_2020}), these dual variables are often explicitly added to the objective. We follow the ``scaled'' ADMM notation used in~\cite{nguyen2024tinympc} to keep the expressions concise and to make the dependence on \(\rho\) more explicit, which is important for our future adaptations.

If we then define the dynamics \(h(\cdot)\) \eqref{eq:dynamics} to be of the time-varying linear affine form (where \(A_k \in \mathbb{R}^{n \times n}\) and \(B_k \in \mathbb{R}^{n \times m}\)):
\begin{align}
x_{k+1} = A_k x_k + B_k u_k,
\end{align}
and the cost $J(\cdot)$ to be of the following quadratic form:
\begin{align}
    J(\cdot) ~&= ~\frac{1}{2}x_N^\top Q_f x_N ~+ ~q_f^\top x_N ~+ \nonumber \\
    ~& \sum_{k=1}^{N-1} \left( \frac{1}{2}x_k^\top Q x_k + q_k^\top x_k + \frac{1}{2}u_k^\top R u_k + r_k^\top u_k \right)
    \label{eq:LQR_cost}
\end{align}
where \(Q \succeq 0\), \(Q_f \succeq 0\), and \(R \succ 0\) are cost matrices, and \(q_k, q_f, r_k\) are cost vectors, then the solution to the primal problem \eqref{ADMM1} becomes the linear-quadratic regulator (LQR) \cite{lewis2012optimal}. This admits an optimal affine feedback policy:
\begin{equation}
u_k^* = -K_k x_k - d_k,\label{eq:lqr_policy}
\end{equation}
where the feedback gain \(K_k\) and feedforward term \(d_k\) are computed via backward recursion of the discrete-time Riccati equation:
\begin{subequations}
\begin{align}
K_k &= (R + B_k^\intercal P_{k+1} B_k)^{-1} (B_k^\intercal P_{k+1} A_k) \label{eq:riccati_gain} \\
d_k &= (R + B_k^\intercal P_{k+1} B_k)^{-1} (B_k^\intercal p_{k+1} + r_k) \label{eq:riccati_ff} \\
P_k &= Q + K_k^\intercal R K_k + (A_k - B_k K_k)^\intercal P_{k+1} (A_k - B_k K_k) \label{eq:riccati_ptg} \\
p_k &= q_k + (A_k - B_k K_k)^\intercal (p_{k+1} - P_{k+1} B_k d_k) + \label{eq:riccati_ltg} \\ 
~& \quad K_k^\intercal (R d_k - r_k), \nonumber
\end{align}
\label{eq:riccati_recursion}%
\end{subequations}
initialized with \(P_N = Q_f\) and \(p_N = q_f\).

\subsection{Caching-Based ADMM for MPC } \label{sec:precomp}

In the case of caching-based methods for MPC, like TinyMPC~\cite{nguyen2024tinympc}, a fixed linearization $(A,B)$ of the system dynamics is used to compute the infinite horizon LQR gain $(K_\infty)$ and cost $(P_\infty)$ as approximations of the time-varying dynamics and previously discussed discrete-time Riccati gain and cost. $A_k, B_k, K_k, P_k$ are therefore replaced with $A, B, K_\infty, P_\infty$ and we can further precompute offline and cache the following matrices:
\begin{equation}
    C_1 = (R + B^\top P_\infty B)^{-1}, 
    \quad 
    C_2 = (A - BK_\infty)^\top.
    \label{C_matrices}
\end{equation}
With $C_1, C_2, K_\infty, P_\infty$ in hand, we can simplify \eqref{eq:riccati_recursion} to only contain the following linear terms:
\begin{equation}\label{eq:fast_riccati}
    \begin{aligned}
        d_k &= C_1(B^\intercal p_{k+1} + r_k), \\
        p_k &= q_k + C_2p_{k+1} - K_\infty ^\intercal r_k.
    \end{aligned}
\end{equation}
This reduces the computational complexity of the Riccati recursion \eqref{eq:riccati_recursion} from $O(n^3)$ to $O(n^2)$.
As such, caching-based MPC leverages a separation of offline caching and online computation for computationally efficient embedded control.

%% file: tex/3_adaptive.tex
Our approach builds on the cached methods discussed in Sec.~\ref{sec:precomp} and integrates adaptive tuning of the ADMM penalty parameter $\rho$, leveraging best practices of normalized residual scaling from~\cite{stellato_osqp_2020}, with efficient updates of the affected cached terms using sensitivity analysis via automatic differentiation. Through the use of first-order Taylor updates to the cached variables, we achieve $O(n^2)$ update complexity rather than the $O(n^3)$ complexity of a full cache computation. Additionally, we introduce a new hyperparameter, $\tau$, that controls how often these updates occur, allowing further balance between computation cost and solution accuracy for efficient embedded implementation.
In the remainder of this section we describe the offline and online steps needed to enact this approach.

\subsection{Offline LQR Sensitivity via Automatic Differentiation}

The ADMM penalty parameter $\rho$ is integrated into the cache through its modification of the cost function hessian:
\begin{equation}
Q_\rho = Q + \rho I_n, \quad R_\rho = R + \rho I_m, \label{eq:QR_wrho}
\end{equation}
which directly impacts the solution of the infinite horizon Riccati terms and the additional cached variables: 
\begin{subequations}
\begin{align}
    K_\infty(\rho) &= (R_\rho + B^T P_\infty B)^{-1} B^T P_\infty A \\
    P_\infty(\rho) &= Q + A^T P_\infty A - \\
    ~& \quad A^T P_\infty B (R_\rho + B^T P_\infty B)^{-1} B^T P_\infty A \nonumber\\
    C_1(\rho) &= (R_\rho + B^T P_\infty B)^{-1} \\
    C_2(\rho) &= (A - B K_\infty)^T
\end{align}
\label{eq:LQR_solution}
\end{subequations}

We leverage automatic differentiation to compute their derivatives with respect to $\rho$, yielding the following sensitivities:
\begin{equation}
    \frac{\partial K_\infty}{\partial \rho}, \quad \frac{\partial P_\infty}{\partial \rho}, \quad \frac{\partial C_1}{\partial \rho}, \quad \frac{\partial C_2}{\partial \rho}.
    \label{eq:sensitivities}
\end{equation}
These terms can then be used in the online step for a reduced computation cache update. Importantly, the inclusion of these additional terms only grows our cache size by a factor of $1.5x$, enabling deployment onto memory constrained embedded systems.

\subsection{Online Adaptive Cache Updates}

To efficiently update the cache online when $\rho$ changes, we leverage a first-order Taylor series approximation of each cached variable. For example, for $K_\infty$ in \eqref{eq:LQR_solution} we have:
\begin{equation}
    K_\infty(\rho + \Delta \rho) \approx K_\infty(\rho) + \frac{\partial K_\infty}{\partial \rho} \Delta \rho,
    \label{eq:updated_lqr_wrho}
\end{equation}
with similar approximations applied to $P_\infty$, $C_1$, and $C_2$. This admits an $O(n^2)$ update online, avoiding the need to resolve the entire LQR problem online for each change in $\rho$, significantly reducing computational overhead and enabling real-time updates in MPC. The only remaining question is then how and when to update the parameter $\rho$.

To do so, we adopt a strategy inspired by the OSQP solver \cite{stellato_osqp_2020}, which uses normalized residuals to balance primal and dual progress. This approach computes scaling factors for the residuals to normalize their magnitudes:
\begin{equation}
    \text{prim\_scaling} = \frac{\|r^{\text{prim}}\|_\infty}{\max\{\|A x\|_\infty, \|z\|_\infty, \phi\}},
    \label{eq:primal_scaling}
\end{equation}
\begin{equation}
    \text{dual\_scaling} = \frac{\|r^{\text{dual}}\|_\infty}{\max\{\|P x\|_\infty, \|A^\top y\|_\infty, \|q\|_\infty, \phi\}},
    \label{eq:dual_scaling}
\end{equation}
where \(\phi = 10^{-8}\) prevents division by zero. These scaling factors reflect the relative magnitudes of the residuals relative to the problem data, enabling a balanced assessment of primal and dual progress.
The penalty parameter \(\rho\) is then updated according to the following rule:
\begin{equation}
    \rho_{k+1} = \rho_k \sqrt{\frac{\text{prim\_scaling}}{\text{dual\_scaling}}}.
    \label{eq:updated_rho}
\end{equation}

\vspace*{-20pt}To balance computational efficiency with adaptation quality, we introduce an update frequency hyperparameter, \(\tau\), which controls how often these updates occur. In our implementation, we set \(\tau = 5\), meaning \(\rho\) is updated every 5 iterations. This update adjusts \(\rho\) to balance the scaled primal and dual residuals, promoting efficient convergence while limiting computational overhead. Following best practice from current state-of-the-art solvers~\cite{stellato_osqp_2020}, to ensure stability, \(\rho_{k+1}\) is clipped to lie within predefined bounds \([\rho_{\min}, \rho_{\max}]\).

Finally, we use the standard convex optimization stopping criteria for ADMM:
\begin{equation}
    \norm{r^{prim}}_{\infty}\leq\epsilon^{prim} ~~\rm{AND}~~  \norm{r^{dual}}_{\infty}\leq\epsilon^{dual},
    \label{eq:stopping_criteria}
\end{equation}
The full online algorithm for our First-Order Adaptive Caching approach, in the context of ADMM based MPC, is outlined in Algorithm~\ref{alg:FOAC}.

%% file: tex/4_experiments.tex
We evaluate our approach in three scenarios of increasing complexity: a near-hover stabilization to establish the improved convergence performance of our approach, a microcontroller timing benchmark on 1000 random motions to demonstrate the embedded feasibility and speed of our approach, and a combination of simulation and hardware deployments of trajectory tracking under wind disturbances to stress-test our resulting controller's robustness. 

\subsection{Near-Hover Stabilization}

We begin with a controlled quadrotor hover scenario that isolates convergence behavior when state deviations remain small. This baseline experiment provides an idealized environment to quantify reduction in iterations without the complexities of hardware constraints or external disturbances. The system accounts for 12-dimensional state vector and 4-dimensional control input with a prediction horizon $N = 10$ and ADMM tolerance $\epsilon^{prim}=\epsilon^{dual}$ set to $10^{-2}$.

\begin{algorithm}[t]
\caption{First-Order Adaptive Caching}
\begin{algorithmic}[1]
\State \textbf{Input:} Initial \(\rho_0\), Cache \(K_{\infty}, P_{\infty}, C_1, C_2\), and their sensitivities~\eqref{eq:sensitivities}, $\epsilon^{prim}$, $\epsilon^{dual}$, $\tau$, $\iota$
\For{$k = 0 \dots \iota$}
    \State Run ADMM steps \eqref{eq:admm_steps} to solve \eqref{eq:admm-MPC} using \eqref{eq:fast_riccati}

    \State Compute residuals via \eqref{eq:residuals}
    \If{\eqref{eq:stopping_criteria} is satisfied} \textbf{break}
    \EndIf
    
    \If{$k \bmod \tau = 0$}
        \State Calculate scaling factors via \eqref{eq:primal_scaling} and \eqref{eq:dual_scaling}
        \State Compute \(\rho_{k+1}\) via \eqref{eq:updated_rho}
        \State Clip \(\rho_{k+1}\) to \([\rho_{\min}, \rho_{\max}]\).
        \State Compute \(\Delta \rho = \rho_{k+1} - \rho_{k}\).
        \State Update cache as in \eqref{eq:updated_lqr_wrho}
    \EndIf
\EndFor
\end{algorithmic}
\label{alg:FOAC}
\end{algorithm}
\begin{figure}
    \centering
    \includegraphics[width=0.9\columnwidth]{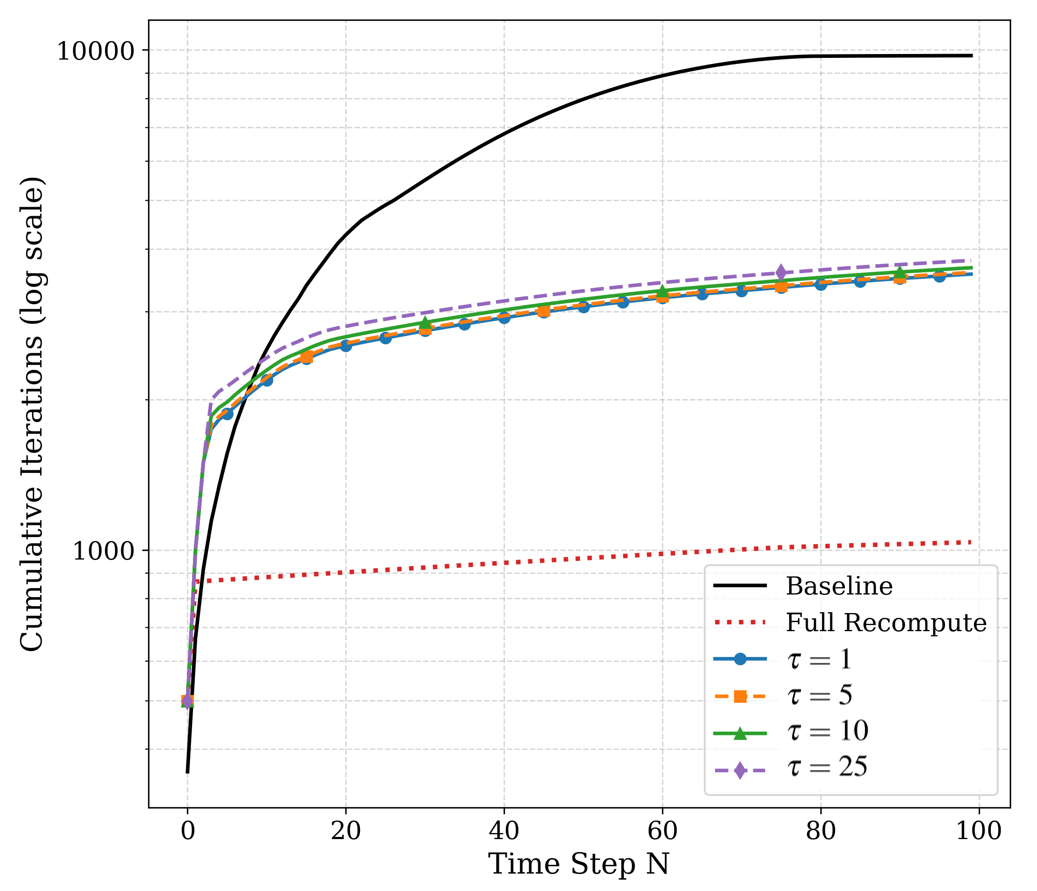}
    \begin{tabular}{lcc}
        \hline
        \textbf{Method} & \textbf{Total Iterations} & \textbf{Speedup} \\
        \hline
        Baseline & 9740 & - \\
        $\tau = 1$ & 3561 & 63.4\% \\
        $\tau = 5$ & 3588 & 63.2\% \\
        $\tau = 10$ & 3667 & 62.4\% \\
        $\tau = 25$ & 3792 & 61.1\% \\
        Full Recompute & 1036 & 89.4\% \\
        \hline
    \end{tabular}
    \caption{Cumulative ADMM iterations across different update frequencies. The baseline method (black) requires the most iterations, while Full Cache Recomputation (red) requires the least number of iterations but is not feasible on the MCU with its $O(n^3)$ cost. First-Order Adaptive Caching reduces iterations by 63.4\%, 63.2\%, 62.4\%, and 61.1\% for updates every 1, 5, 10, and 25 steps respectively and approaches 70\% of the full recompute performance at $O(n^2)$ cost.}
    \label{fig:iteration_comparison}
    \vspace{-15pt}
\end{figure}

Our comparison spans three approaches: 1) Fixed $\rho$ using manually tuned static parameters through an exhaustive grid search; 2) Adaptive $\rho$ leveraging our normalized residual optimization; and 3) Full Cache Recomputation representing an idealized but computationally prohibitive baseline. To determine the empirically stable range for our hover experiments, we systematically tested fixed $\rho$ values and found that values between 60 and 100 produced the best results. This finding aligns with ADMM theory, where higher $\rho$ values tend to improve primal residual convergence while potentially slowing dual residual convergence \cite{boyd2011distributed}. For hover stabilization, where state deviations remain small and constraint satisfaction is critical, a higher penalty parameter ($\rho = 85.0$) effectively prioritizes constraint feasibility while maintaining reasonable numerical conditioning to avoid numerical instability.

Fig.~\ref{fig:iteration_comparison} shows the cumulative ADMM iterations across 100 MPC steps. First-Order Adaptive Caching consistently outperforms the fixed approach, reducing total iterations by 63.4\%, 63.2\%, 62.4\%, and 61.1\% for update frequencies of 1, 5, 10, and 25 steps respectively. Full Cache Recomputation achieves the greatest reduction (90.8\%), but is computationally prohibitive for embedded implementation. Notably, our method achieves 70\% of the full recomputation performance at an order of complexity less compute and maintains most of its performance advantage even when updating $\rho$ less frequently, indicating robustness to update frequency and suitability for resource-constrained systems.

\begin{figure}[t]
    \centering
    \vspace{5pt}
    \includegraphics[width=1.0\linewidth]{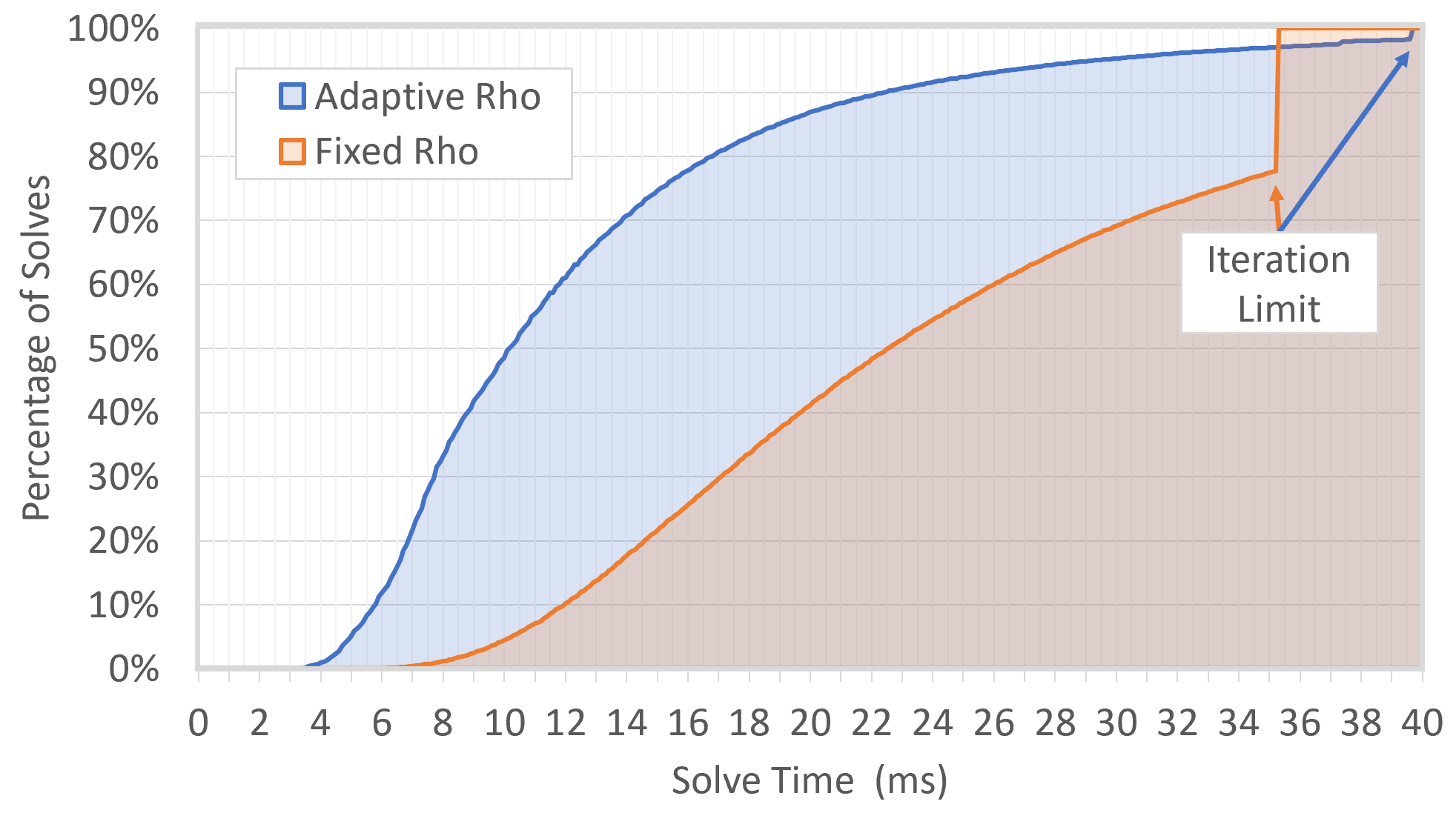}
    \caption{Cumulative distribution function (CDF) of solve times with $\rho = 85.0$ across 1000 random goal states and inputs and 100 random controllable systems with quadrotor dimensions. We find that the adaptive method solves at much faster rates due to our low-overhead cache adaptation.} 
    \label{fig:cdf_plot}
    \vspace{-10pt}
\end{figure}

\begin{table}[t]
\centering
\vspace{5pt}
\caption{Solver Performance Statistics with $\rho = 85.0$ across 1000 random goal states and inputs, showing the performance gains and low overhead of our approach.}
\label{tab:solver_stats}
\begin{tabular}{lcc}
\hline
\textbf{Metric} & \textbf{Fixed $\rho$} & \textbf{Adaptive $\rho$} \\
\hline
Mean Solve Time (ms) & 23.459 & \textbf{12.488} \\
Median Solve Time (ms) & 22.546 & \textbf{10.168} \\
95th Percentile Time (ms) & 35.232 & \textbf{29.320} \\
Standard Deviation (ms) & 8.834 & \textbf{7.580} \\
\hline
\end{tabular}
\vspace{-15pt}
\end{table}

\subsection{Microcontroller Timing Benchmarks}

Next we test our approach on a physical microcontroller unit (MCU) to evaluate its real-time performance across a wider range of problems. In particular, we seek to determine whether the adaptation overhead outweighs our improved convergence rates. We run our tests on a Teensy 4.1 microcontroller (IMXRT1062) which has a 600 MHz ARM Cortex-M7 with 1024K RAM.

We benchmark our adaptive approach against the fixed approach using 100 randomly generated controllable systems with 12 states and 4 inputs, matching the dimensionality of typical quadrotor models with $\rho = 85.0$. For each system, we solve 1000 MPC problems with randomly generated reference trajectories, where both the state ($X_{\text{ref}}$) and control ($U_{\text{ref}}$) trajectories are randomly generated for each trial. Unlike the hover tests, which focused on a specific control scenario, these random problems represent a comprehensive sampling of potential operating conditions, including hover states, dynamic trajectories, and various transitional states. This broader test suite provides a much more thorough evaluation of our method's robustness and adaptability. Fig.~\ref{fig:cdf_plot} shows the cumulative distribution function (CDF) of solve times for both fixed and adaptive approaches and Table~\ref{tab:solver_stats} summarizes the key performance metrics from our benchmark.

Overall, we find that our adaptive approach only incurs a 5.4\% average computational overhead while significantly reducing the number of solver iterations across the majority of the problems, and thus resulting in much faster solve times. In fact, we achieve a 46.8\% reduction in mean solve time (12.49\,ms vs 23.46\,ms), a 54.9\% reduction in median solve time (10.17\,ms vs 22.55\,ms), and a 14.2\% reduction in solve time standard deviation (7.58\,ms vs 8.83\,ms). Our adaptive approach solves 97.9\% of problems under the 500 iteration limit, compared to only 77.8\% for the fixed approach. Additionally, only 9.8\% of adaptive solves were slower than the fixed baseline, and our fastest solve was 29.3\% faster, while the worst-case solve was only 12.5\% slower than the worst fixed case, highlighting a substantial improvement in both typical and edge-case solver performance.

\subsection{Figure-Eight Trajectory Tracking with Wind}

While our hover experiments and MCU benchmarking demonstrate improved convergence speed and robustness, this section evaluates our method's performance in dynamic scenarios with external disturbances to evaluate the real-world applicability of our approach. We test a quadrotor executing a high-speed figure-eight trajectory while subjected to wind disturbances of varying magnitudes in both simulation and on a 27g Crazyflie 2.1+~\cite{giernacki2017crazyflie}.

For this more challenging scenario, we set $N = 15$ and set $\epsilon^{prim}=\epsilon^{dual}=10^{-3}$ as used in the original TinyMPC \cite{nguyen2024tinympc} codebase.
Based on our hover experiment results, we implement $\rho$ updates every 5 iterations, as this frequency provides a good balance between adaptation benefits and computational efficiency. We also limit the maximum ADMM iterations to 10, matching TinyMPC's implementation \cite{nguyen2024tinympc}, to ensure compatibility with the Crazyflie's MCU constraints.

\subsubsection{Simulation Experiment}

We initialize $\rho=5.0$, the value specified in the TinyMPC codebase \cite{nguyen2024tinympc} for figure-eight maneuvers. According to \cite{nguyen2024tinympc}, this value was determined through offline optimization and represents an optimal parameter for this scenario. This lower value is theoretically justified for dynamic trajectories with external disturbances, as it allows greater flexibility in the dual variable updates while preventing over-penalization that could lead to numerical issues when the system is pushed away from its nominal trajectory. The dynamic nature of this task benefits from allowing some constraint violation during transients, which a lower $\rho$ facilitates. The contrast between the hover scenario ($\rho = 85.0$) and figure-eight trajectory ($\rho = 5.0$) highlights how ADMM penalty parameters typically need scenario-specific tuning, a challenge our adaptive method aims to address. The quadrotor executes multiple laps of the reference trajectory while experiencing wind disturbances modeled as:
\begin{equation}
    \mathbf{w}(t) = m \begin{bmatrix}
        \cos(\theta) \\
        \sin(\theta) \\
        \eta
    \end{bmatrix},
\end{equation}
where $m=25.5~[m/s^2]$ (nominally) represents the wind acceleration magnitude, $\theta \sim \mathcal{U}(0, 2\pi)$ provides the horizontal wind direction, and $\eta \sim \mathcal{U}(-0.3, 0.3)$ introduces vertical disturbance components from a uniform distribution.

Table \ref{tab:merged_performance} quantifies our experimental results across both standard and wind disturbance conditions. Under standard conditions without wind, our adaptive method requires 71.9\% fewer iterations while maintaining comparable tracking performance (only 8.6\% higher average error). This dramatic reduction in computational load with minimal performance impact demonstrates the efficiency of our First-Order Adaptive Caching approach even in nominal conditions.

The most revealing results emerge under wind disturbance conditions. The adaptive method achieves a 20.3\% reduction in average tracking error and a substantial 23.0\% reduction in maximum error compared to the fixed approach. This improved robustness comes with identical iteration counts, indicating that our method allocates computational effort more intelligently when facing disturbances. As shown in Figure~\ref{fig:traj_compare}, the adaptive approach demonstrates superior trajectory tracking capabilities, particularly under external disturbances, where it more closely follows the reference trajectory compared to the fixed method.

Our findings indicate that adaptive $\rho$ tuning becomes increasingly valuable as conditions become more challenging. While both approaches perform similarly in ideal conditions, our adaptive method demonstrates significantly better disturbance rejection capabilities when facing unpredictable wind—precisely the scenario where robust control is most critical for real-world deployment.

\subsubsection{Hardware Deployment}
To test robustness in the real world, we deploy our approach on a Crazyflie 2.1+, a 27 gram quadrotor with a Cortex M4 MCU serving as its main application, offering 168MHz, 192kb SRAM, and 1Mb flash. As shown in Figure~\ref{fig:hero} our First-Order Adaptive Method successfully navigates the figure-eight trajectory in the presence of wind disturbances from a fan, closing the sim-to-real gap. Our demos can be viewed at: \href{https://github.com/A2R-Lab/FoAC}{\texttt{https://github.com/A2R-Lab/FoAC}}.

%% file: tex/5_conclusion.tex
We introduced First-Order Adaptive Caching, a novel approach that addresses a fundamental limitation in existing caching-based MPC solvers for resource-constrained systems. By extending TinyMPC \cite{nguyen2024tinympc} with dynamic cache updates via first-order Taylor expansions and implementing an adaptive tuning rule for the ADMM penalty parameter \(\rho\), we achieved significant improvements in computational efficiency while maintaining robustness across diverse operating conditions.
This performance gain is maintained across both stable hover conditions and challenging dynamic scenarios with wind disturbances, where our adaptive approach delivers superior trajectory tracking accuracy. 

There are many promising directions for future work. For example, although extensive empirical evidence supports the stability of our approach under dynamic cache updates and adaptive tuning, establishing rigorous theoretical guarantees would further enhance confidence for safety-critical applications.
Furthermore, machine learning-based approaches  could potentially tune \(\rho\) even more effectively while maintaining computational efficiency. 

\begin{figure}[t]
    \centering
    \includegraphics[width=\columnwidth]{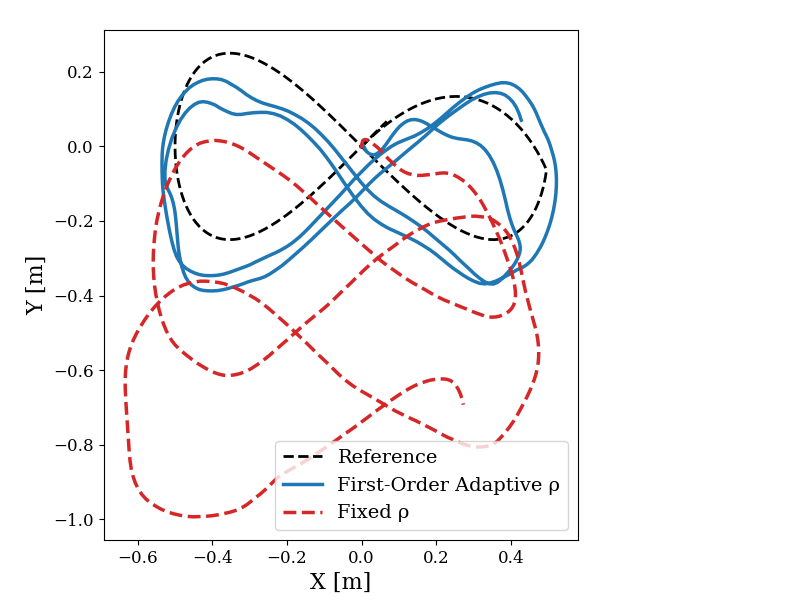}
    \caption{Tracking performance comparison between fixed and adaptive penalty parameter methods at the highest wind magnitude maintaining numerical stability in simulation. The adaptive method (blue) demonstrates more precise tracking at curve extremes versus the fixed method (red).}
    \label{fig:traj_compare}
    \vspace{-5pt}
\end{figure}

\begin{table}[t]
    \centering
    \caption{Figure-Eight Metrics under Standard and Wind Disturbance Conditions for both the (F)ixed and (A)daptive $\rho$ Methods}
    \label{tab:merged_performance}
    \begin{tabular}{lcc|cc}
    \hline
    \multirow{2}{*}{\textbf{Metric}} & \multicolumn{2}{c}{\textbf{Standard}} & \multicolumn{2}{c}{\textbf{Wind}} \\
    & \textbf{F} & \textbf{A} & \textbf{F} & \textbf{A} \\
    \hline
    Avg. L2 Error (m) & \textbf{0.03} & 0.03 & 0.74 & \textbf{0.59} \\
    Max. L2 Error (m) & \textbf{0.09} & 0.10 & 1.04 & \textbf{0.80} \\
    Avg. Iterations/Step & 7.87 & \textbf{2.21} & 8.98 & \textbf{8.97} \\
    Total Iterations & 3148 & \textbf{886} & 3591 & \textbf{3590} \\
    \hline
    \end{tabular}
    \vspace{-10pt}
\end{table}